# Özgeçmişlerde varlık isimlerinin tanınması
# Named entity recognition in resumes

Ege Kesim, Aysu Deliahmetoglu (Intern)
ege.kesim1@huawei.com, aysudeliahmetoglu@gmail.com
Huawei Turkey Research and Development Center, Istanbul

*Özetçe*—Varlık İsmi Tanıma (VİT) çeşitli belge ve metinlerden isim, tarih gibi çeşitli bilgilerin çıkarılmasında kullanılmaktadır. Özgeçmişlerden kişilerin eğitim seviyeleri ve daha önceki çalışma deneyimi ile ilgili bilgilerin çıkarılması özgeçmişlerin filtrelenebilmesi için önem arz etmektedir. Her özgeçmişteki bilgilerin şirketlerin sistemlerine manuel olarak girileceği göz önüne alınınca bunun otomatik bir şekilde yapılması şirketlere vakit kazancı sağlamaktadır. Bu çalışmada, bilişim alanındaki özgeçmişler özelinde derin öğrenme tabanlı yarı-otomatik bir varlık ismi tanıma sistemi gerçekleştirilmiştir. İlk olarak bilişim sektöründeki beş farklı alanda başvuran adayların özgeçmişleri etiketlenmiştir. Daha önceden eğitilmiş altı farklı dönüştürücü tabanlı model varlık ismi tanıma için etiketlenen veriye adapte edilmiştir. Bu modeller farklı doğal dil işleme problemlerinde kullanılan popüler modeller arasından seçilmiştir. Elde edilen sistem şehir, tarih, diploma, diploma bölümü, iş unvanı, lisan, ülke ve yetenek olmak üzere sekiz farklı varlık türünü tanıyabilmektedir. Yapılan deneylerde kullanılan modeller mikro, makro ve ağırlıklı F1 skorları kullanılarak karşılaştırılmış ve yöntemlerin başarısı değerlendirilmiştir. Bu skorlar dikkate alındığında test kümesi için en iyi mikro ve ağırlıklı F1 skorunu RoBERTa, en iyi makro F1 skorunu Electra modeli elde etmektedir.

*Anahtar Kelimeler* — *Varlık ismi tanıma, doğal dil işleme, insan kaynakları, BERT, dönüştürücü tabanlı dil modelleri*

*Abstract*— Named entity recognition (NER) is used to extract information from various documents and texts such as names and dates. It is important to extract education and work experience information from resumes in order to filter them. Considering the fact that all information in a resume has to be entered to the company's system manually, automatizing this process will save time of the companies. In this study, a deep learning-based semi-automatic named entity recognition system has been implemented with a focus on resumes in the field of IT. Firstly, resumes of employees from five different IT related fields has been annotated. Six transformer based pre-trained models have been adapted to named entity recognition problem using the annotated data. These models have been selected among popular models in the natural language processing field. The obtained system can recognize eight different entity types which are city, date, degree, diploma major, job title, language, country and skill. Models used in the experiments are compared using micro, macro and weighted F1 scores and the performance of the methods was evaluated. Taking these scores into account for test set the best micro and weighted F1 score is obtained by RoBERTa and the best macro F1 score is obtained by Electra model.

*Keywords* — *Named entity recognition, natural language processing, human resources, BERT, transformer based language models*

I. GİRİŞ

Varlık ismi tanıma metinlerde geçen adres, şahıs ve şirket isimleri gibi çeşitli varlıklara ait isimlerin bulunması problemidir. Varlık ismi tanıma doğal dil işlemede sıklıkla kullanılmakta olup özellikle belge ve dokümanlarda geçen çeşitli kelimelerin bulunmasında kullanılmaktadır.

BILSTM-CRF modelleri varlık ismi tanıma probleminde sıklıkla kullanılmaktadır. BILSTM-CRF modeli varlık ismi tanıma probleminde başarılı sonuçlar vermiştir [1,2]. Ayrıca BERT modelinin dondurulup sadece bir kelime temsil modeli olarak kullanıldığı [2] ya da ince ayar yapılarak varlık ismi tanımaya adapte edildiği çalışmalar da mevcuttur [3,4].

Varlık ismi tanıma insan kaynakları sektöründe de kullanılmaktadır. Şirketlerin çok sayıda iş başvurusu almasından dolayı başvuran adayların özgeçmişlerinin işlenmesi ve saklanması önemli problemlerdir. Varlık ismi tanıma kullanılarak özgeçmişlerdeki önemli bilgiler çıkarıldıktan sonra bu bilgilere göre filtreleme yapılabilir ya da özgeçmişin tamamı yerine sadece bu gerekli bilgiler sistemlere aktarılıp saklanabilir.

Varlık ismi tanıma popüler bir problem olmasına rağmen açık kaynaklı özgeçmiş veri kümelerinin kısıtlı olması sebebiyle bu alandaki varlık ismi tanıma çalışmaları kısıtlıdır. Çoğu çalışmada internet üzerinden özgeçmişler toplanıp kullanılmaktadır. Gaur ve diğerleri [5] çalışmalarında özgeçmişlerin eğitim bölümü kullanılarak 3 farklı varlık türünü tanıyabilen CNN-BILSTM tabanlı bir model geliştirmiştir. Zu ve Wang [6] 5000 özgeçmiş kullanarak 19 farklı varlık türünü tanıyabilen CNN-BILSTM-CRF tabanlı bir model geliştirmiş ve bu modelin sadece BILSTM-CRF kullanmaktan daha iyi olduğunu göstermiştir.

Bunların dışında literatürde Çince özgeçmişlerden oluşan veri kümesi bulunmasından dolayı Çince özgeçmişlerle yapılan çalışmalar da mevcuttur [7,8].

Bu çalışmada İngilizce özgeçmişler kullanılarak varlık ismi tanıma yapılmıştır. Kuruluşun İK veritabanlarında tutulan özgeçmiş verilerinden seçilen örnekler anonimleştirilmiş ve genelleştirilmiş; IT alanında özgün bir veri kümesi oluşturulmuştur. Bilişim sektöründe beş farklı alanda çalışan kişilerin özgeçmişlerinden oluşan veriler, yarı-otomatik bir şekilde sekiz farklı varlık türünü kapsayacak şekilde etiketlenmiştir. Bu varlık isim türleri şehir, tarih, diploma, diploma bölümü, iş unvanı, lisan, ülke ve yetenekten oluşmaktadır. Ayrıca etiketleme süreci boyunca çeşitli yapay zeka modelleri kullanılarak etiketleme süreci hızlandırılmıştır. Etiketlenen bu veriler kullanılarak BERT ve Roberta gibi daha önceden eğitilmiş altı farklı dil modeli varlık ismi tanıma problemine adapte edilmiş ve performansları karşılaştırılmıştır. Modellerin performans karşılaştırmaları için makro, mikro ve ağırlıklı F1 skorları kullanılmıştır.

## II. YÖNTEM

### A. Verilerin Etiketlenmesi

Açık kaynaklı olan veri kümelerinin çoğu hem etiketlenmemiş hem de çok az sayıda özgeçmişten oluşmaktadır. Bu çalışmada bilişim sektöründe beş farklı alanda çalışan kişilerin özgeçmişleri etiketlenmiştir. Verilerin etiketlenmesi için açık kaynaklı bir araç olan Label Studio[1] kullanılmıştır.

Şekil. 1. Etiketlenmiş bir veri örneği (Eğitim bölümü için)

Şekil. 2. Etiketlenmiş bir veri örneği (Deneyim bölümü için)

Özgeçmişlerin içerisinde çok fazla varlık ismi olmasından dolayı etiketlenmesi zahmetlidir. Bir özgeçmişte eğer kişinin sadece lisans eğitimi varsa tek bir diploma bilgisi bulunmaktadır. Ancak aynı özgeçmişte bazen onlarca yetenek bulunabilmektedir (Örneğin: Python, Git, vb.). Bazı varlık isimlerinin çok az geçmesi gözden kaçırmaya, çok sık geçmesi ise etiketlemenin zahmetli olmasına sebebiyet vermektedir. Bu problemi çözmek için etiketleme yarı-otomatik bir şekilde yapılmıştır. Yarı otomatik etiketleme dört aşamadan oluşmaktadır. İlk aşamada veri kümesinin küçük bir bölümü bazı hazır varlık ismi tanıma modelleri kullanılarak otomatik olarak etiketlenmiştir. İkinci aşamada ise bu etiketler bir insan tarafından kontrol edilerek düzeltilmiştir.

Üçüncü aşamada bu veriler kullanılarak bir varlık ismi modeli eğitilmiştir. Böylece performansı iyi olmasa da bulmak istediğimiz tüm varlık isimlerini tanıyabilen bir sistem elde edilmiştir. Bu yeni sistem ile aynı ilk aşamada olduğu gibi verinin tamamı otomatik olarak etiketlenmiştir. Son aşamada ise tüm veriler bir insan tarafından tekrar kontrol edilmiş ve düzeltilmiştir. Tüm bu aşamalardan sonra veri kümesinin son hali elde edilmiştir. İlerleyen bölümlerdeki model performansları veri setinin son hali ile eğitilen modellerin performanslarıdır. Ara aşamalardaki modeller sadece etiketlemeyi kolaylaştırmak amacıyla kullanılmıştır.

TABLO I. VERİ KÜMELERİ ARASINDAKİ ETİKET DAĞILIMI

| Varlık Türü | Eğitim | Doğrulama | Test |
|---|---|---|---|
| Şehir | 463 | 78 | 90 |
| Tarih | 975 | 192 | 234 |
| Diploma | 176 | 45 | 33 |
| Diploma bölümü | 291 | 63 | 65 |
| İş unvanı | 641 | 125 | 147 |
| Lisan | 393 | 101 | 85 |
| Ülke | 329 | 65 | 56 |
| Yetenek | 2885 | 610 | 618 |

### B. Veri Kümesi

Veri kümesi toplamda 280 adet anonimleştirilmiş ve genelleştirilmiş özgeçmişten oluşmaktadır. Her bir özgeçmiş kişisel, eğitim, iş deneyimi, lisan ve yetenek olmak üzere beş ayrı bölümden oluşmaktadır. Veri kümesi içerisinde toplamda 1430 adet bölüm bulunmaktadır. Her bir bölüm ayrı ayrı etiketlenmiş ve eğitilen modellere ayrı ayrı girdiler olarak verilmiştir. Kısacası modeller aynı anda tüm özgeçmiş yerine tek bir bölümü almaktadır. Bölümler kişi bazlı olarak eğitim, doğrulama ve test olarak üç kümeye ayrılmıştır. Kümelerin oranları yaklaşık olarak 70/15/15 şeklindedir. Veriler bu gruplara dağıtılırken hem etiket sayılarının hem de başvurulan işlere göre kümeler içerisindeki yüzdesel dağılımlar birbirlerine yakın olacak şekilde seçilmiştir. Kümeler arası etiket dağılımı Tablo I'de , kümeler arası alan dağılımı ise Tablo II'de görülebilir.

---
[1] https://labelstud.io/

TABLO II. VERİ KÜMELERİ ARASINDAKİ ALAN DAĞILIMI

| Alan | Eğitim | Doğrulama | Test |
|---|---|---|---|
| Ai Engineer | 54 | 12 | 12 |
| Android Developer | 37 | 8 | 8 |
| Developer Partner Engineer | 40 | 9 | 9 |
| Developer Support Engineer | 32 | 7 | 8 |
| Senior Software Developer | 35 | 7 | 8 |

*C. Karşılaştırılan Modeller*

Bu çalışmada dönüştürücü tabanlı önceden eğitilmiş farklı dil modelleri varlık ismi tanıma problemine adapte edilmiştir. Bu bölümde kullanılan modeller özetlenmektedir.

*1) BERT:* BERT [9] modeli iki aşamalı şekilde eğitilmiş çift yönlü dönüştürücü tabanlı bir dil modelidir. İlk eğitim aşamasında etiketlenmemiş veriler maskelenmiş dil modelleme ve sonraki cümlenin tahmini görevleriyle eğitilir. Daha sonra farklı dil problemleri için modele ek çıktı katmanları eklenerek modelin varlık ismi tanıma, metin sınıflandırması, duygu analizi gibi problemler için eğitimi sağlanır. Bu çalışmada önceden eğitilmiş iki farklı BERT varyasyonu kullanılmıştır. Birinci versiyon bert-base-cased[2] ikincisi ise bert-base-uncased[3] modelleridir. Bu modellerden bert-cased büyük-küçük harf kullanımına karşı duyarlıyken bert-uncased modeli büyük-küçük harf ayrımı yapmamaktadır.

*2) DistilBert:* DistilBert [10] modeli BERT modelinin damıtılması ile oluşturulmuş BERT benzeri bir yapıya sahip ancak parametre sayısı açısından %40 daha küçük ve %60 daha hızlı çalışan bir dil modelidir. Çeşitli katmanlar modelden çıkartılarak parametre sayısı azaltılmıştır. Bu model BERTten küçük olmasına rağmen ona yakın bir performansa sahiptir. Büyük-küçük harf ayrımı yapmayan distilbert-uncased[4] modeli deneylerde kullanılmıştır.

*3) RoBERTa:* RoBERTa [11] modeli BERT ile aynı yapıya sahip ancak daha optimize bir eğitime tabi tutulmuş dil modelidir. BERT modelinden farklı olarak statik maskelenmiş dil modelleme yerine dinamik bir mekanizma kullanılarak eğitilmiştir. Böylece BERTteki gibi aynı şekildeki maskelenmiş girdiler yerine her adımda aynı girdileri farklı maskeleme şekillerinde görmektedir. Ayrıca BERT modelinde kullanılan sonraki cümle tahmini RoBERTa'nın eğitiminde kullanılmamaktadır. RoBERTa-base5 modeli bu çalışmada kullanışmıştır.

*4) Xlm-RoBERTa:* XLM-RoBERTa [12] dönüştürücü tabanlı 100 farklı dil üzerinde eğitilmiş bir dil modelidir. Bu model RoBERTada kullanılan tekniklerin XLM modeline uygulanması ile elde edilmiştir. Model 2.5 Terabaytlık filtrelenmiş CommonCrawl veri kümesi kullanılarak eğitilmiştir. Yapılan deneylerde XLM-RoBERTa-base6 modeli kullanılmıştır.

*5) ELECTRA:* ELECTRA [13] maskelenmiş dil modellemesi için BERT gibi modellerden daha farklı bir yaklaşım geliştirerek üretken çekişmeli ağlarını kullanmıştır. Bu yaklaşımda girdiler içerisindeki dizgecikler maskelenmek yerine bir üreteci model tarafından rastgele sahte dizgecikler ile değiştirilir. Ayırıcı olarak isimlendirilen ikinci bir model ise dizgeciklerin sahte yada gerçek olduğunu ayırt etmek üzere eğitilmiştir. Önceden eğitilmiş bu ayırıcı ağ daha sonra ince ayar yaparak farklı görevlere adapte edilebilir. Bu çalışmada ELECTRA-base[7] modeli kullanılmıştır.

III. DENEYLER

Modeller Python dili ile Spacy[8] kütüphanesi kullanılarak eğitilmiştir. Önceden eğitilmiş modellerin ağırlıkları Huggingface'ten alınmıştır. Modellerin eğitimi için bekleme değeri beş devir (epoch) olarak seçilmiştir. Beş devirden sonra performansın artmaması halinde eğitim durdurulmuştur.

Tablo III'de modellerin mikro, makro ve ağırlıklı F1 skorları verilmektedir. Bu skorlar karşılaştırıldığında en iyi makro F1 skorunu Electra modeli elde etmektedir. Mikro ve ağırlıklı F1 skorlarında ise en başarılı skoru RoBERTa modeli elde etmiştir. DistilBERT damıtılmış bir model olduğundan diğer modellere göre daha kötü performans elde etmesi beklenmektedir. Ancak üç metrikten ikisinde en kötü skoru almasına rağmen diğer modellere yakın performans elde etmiştir. İşlem hızının ve hacminin öncelikli olduğu durumlarda performanstan ödün verilerek bu model tercih edilebilir.

TABLO III. VARLIK İSMİ TANIMA PERFORMANS KARŞILAŞTIRMA TABLOSU

| Model İsmi | Mikro F1 | Makro F1 | Ağırlıklı F1 |
|---|---|---|---|
| BERT (cased) | 86.95 | 87.91 | 87.03 |
| BERT (uncased) | 88.4 | 89.66 | 88.39 |
| DistilBERT | 86.58 | 88.47 | 86.53 |
| ELECTRA | 88.56 | **90.19** | 88.55 |
| RoBERTa | **89.58** | 89.28 | **89.63** |
| XLM-RoBERTa | 87.08 | 87.5 | 87.06 |

---

[2] https://huggingface.co/bert-base-cased
[3] https://huggingface.co/bert-base-uncased
[4] https://huggingface.co/distilbert-base-uncased
[5] https://huggingface.co/roberta-base
[6] https://huggingface.co/xlm-roberta-base
[7] https://huggingface.co/google/electra-base-discriminator
[8] https://spacy.io/

Tablo IV ve Tablo V'de Electra ve RoBERTa modellerinin her bir varlık türü için keskinlik, hassasiyet ve F1 metriklerine göre performansları verilmiştir. ELECTRA modelinde en düşük F1 skoru iş unvanı için 80, diğer varlık türlerinden 85 ve üzerindedir. RoBERTa modelinde ise en düşük F1 skoru 72 ile diploma varlık türü için elde edilmiş, geri kalan varlık türlerinde 86 üzerinde F1 skoru elde edilmiştir. Her iki model için de en yüksek performans alınan üç varlık türü lisan, ülke ve şehirdir.

TABLO IV. ELECTRA MODELİ İÇİN VARLIK TÜRÜ KARŞILAŞTIRMA TABLOSU

| Varlık Türü | Keskinlik | Hassasiyet | F1 |
|---|---|---|---|
| Şehir | 94.57 | 96.67 | 95.60 |
| Tarih | 81.30 | 91.03 | 85.89 |
| Diploma | 85.29 | 87.88 | 86.57 |
| Diploma bölümü | 90.77 | 90.77 | 90.77 |
| İş unvanı | 82.86 | 78.91 | 80.84 |
| Lisan | 97.70 | 100.00 | 98.84 |
| Ülke | 93.10 | 96.43 | 94.74 |
| Yetenek | 87.70 | 88.83 | 88.26 |

TABLO V. ROBERTA MODELİ İÇİN VARLIK TÜRÜ KARŞILAŞTIRMA TABLOSU

| Varlık Türü | Keskinlik | Hassasiyet | F1 |
|---|---|---|---|
| Şehir | 96.67 | 96.67 | 96.67 |
| Tarih | 83.52 | 95.3 | 89.02 |
| Diploma | 64.29 | 81.82 | 72.0 |
| Diploma bölümü | 88.71 | 84.62 | 86.61 |
| İş unvanı | 89.29 | 85.03 | 87.11 |
| Lisan | 96.59 | 100 | 98.27 |
| Ülke | 94.74 | 96.43 | 95.58 |
| Yetenek | 87.25 | 90.78 | 88.98 |

Her iki model varlık türü bazında karşılaştırıldığında RoBERTa'nın çoğu varlık türünde daha iyi performans gösterdiği görülebilir. Ağırlıklı F1 skorunun anlaşılması için Tablo I kullanılarak test kümesi için varlık türlerinin dağılımı incelebilir. RoBERTa diploma varlık türü için ELECTRA modeline göre çok daha düşük performans vermesine rağmen en az örnek sayısını bulundurmasından dolayı ağırlıklı ortalamada etkisi düşük olmuştur. ELECTRA modelinin RoBERTa modelinden iyi olduğu varlık türleri çoğunlukla daha öz örnek bulunan varlık türleridir.

## IV. SONUÇ

Bu çalışmada, bilişim alanındaki özgeçmişlere odaklı bir varlık ismi tanıma sistemi elde edilmiştir. Literatürde sıklıkla kullanılan altı farklı önceden eğitilmiş dil modeli ince ayarlama ile özgeçmişlerde varlık ismi tanıma için adapte edilmiştir. Modeller mikro, makro ve ağırlık F1 metrikleri kullanılarak karşılaştırılmıştır. Yapılan deneyler sonucunda RoBERTa mikro ve ağırlıklı F1 metriklerinde en iyi, ELECTRA modeli ise makro F1 metriğinde en iyi performansı alan model olmuştur. Ayrıca bu modellere göre daha hızlı ve küçük bir model olan DistilBERT modelinin çok az bir performans kaybı ile kullanılabileceği gözlemlenmiştir. Bu çalışmada bilişim sektöründeki beş farklı alanda çalışan kişilerin özgeçmişleri kullanılmıştır. Elde edilen sistem sekiz farklı varlık isim türünü tanıyabilmektedir. Yürütülen bu özgün çalışmanın sonuçları ışığında, veri kümesindeki örneklem sayısı arttırılarak, farklı iş kollarından gelen özgeçmişlerle çeşitlendirilerek geliştirilmeye açık devam çalışmaları öngörülmektedir. Ayrıca, derin öğrenme tabanlı yaklaşımların, şirket taleplerini potansiyel adaylarla birleştirmedeki muazzam avantajlarından yararlanarak işverenler için zaman tasarrufu sağlayacak bir araç tasarımı planlanmaktadır.

## KAYNAKLAR

[1] Jie, Z., and Lu, W., "Dependency-guided LSTM-CRF for named entity recognition." arXiv preprint arXiv:1909.10148, 2019.

[2] Tanrısever, Özer, et al. "Named Entity Recognition for Defense Industry." 2022 30th Signal Processing and Communications Applications Conference (SIU). IEEE, 2022.

[3] Souza, Fábio, Rodrigo Nogueira, and Roberto Lotufo. "Portuguese named entity recognition using BERT-CRF." arXiv preprint arXiv:1909.10649 (2019).

[4] Darji, Harshil, Jelena Mitrović, and Michael Granitzer. "German BERT Model for Legal Named Entity Recognition." arXiv preprint arXiv:2303.05388 (2023).

[5] Gaur, Bodhvi, et al. "Semi-supervised deep learning based named entity recognition model to parse education section of resumes." Neural Computing and Applications 33 (2021): 5705-5718.

[6] Zu, Shicheng, and Xiulai Wang. "Resume information extraction with a novel text block segmentation algorithm." Int J Nat Lang Comput 8 (2019): 29-48.

[7] Zhu, Peng, et al. "Improving Chinese named entity recognition by large-scale syntactic dependency graph." IEEE/ACM Transactions on Audio, Speech, and Language Processing 30 (2022): 979-991.

[8] Han, Xiaokai, et al. "Multi-Feature Fusion Transformer for Chinese Named Entity Recognition." 2022 41st Chinese Control Conference (CCC). IEEE, 2022.

[9] Devlin, Jacob, et al. "Bert: Pre-training of deep bidirectional transformers for language understanding." arXiv preprint arXiv:1810.04805 (2018).

[10] Sanh, Victor, et al. "DistilBERT, a distilled version of BERT: smaller, faster, cheaper and lighter." arXiv preprint arXiv:1910.01108 (2019).

[11] Liu, Yinhan, et al. "Roberta: A robustly optimized bert pretraining approach." arXiv preprint arXiv:1907.11692 (2019).

[12] Conneau, Alexis, et al. "Unsupervised cross-lingual representation learning at scale." arXiv preprint arXiv:1911.02116 (2019).

[13] Clark, Kevin, et al. "Electra: Pre-training text encoders as discriminators rather than generators." arXiv preprint arXiv:2003.10555 (2020).